\documentclass{article}

\usepackage{arxiv}
\usepackage[utf8]{inputenc} 
\usepackage[T1]{fontenc}    
\usepackage{hyperref}       
\usepackage{url}            
\usepackage{booktabs}       
\usepackage{amsfonts}       
\usepackage{nicefrac}       
\usepackage{microtype}      
\usepackage{lipsum}
\usepackage{fancyhdr}       
\usepackage{array}
\usepackage{cite}  
\usepackage{amsmath}
\usepackage{amsfonts}
\usepackage{dcolumn}
\usepackage{graphicx} 

\newcommand{\newsec}[1]{%
    \vspace{\baselineskip}  
    \noindent 
    \textbf {#1} 
    \vspace{\baselineskip}  
}

\newcolumntype{P}[1]{>{\centering\arraybackslash}p{#1}}

\title{Fast Mutual Information Computation for Large Binary Datasets}

\author{
  Andre O. Falcao\\
  BioISI - Biosystems and Integrative Sciences. \\
  Faculdade de Ciências - Universidade de Lisboa \\
  Lisboa\\
  \texttt{aofalcao@ciencias.ulisboa.pt} \\
}

\begin{document}

\maketitle

\begin{abstract}
Mutual Information (MI) is a powerful statistical measure that quantifies shared information between random variables, particularly valuable in high-dimensional data analysis across fields like genomics, natural language processing, and network science. However, computing MI becomes computationally prohibitive for large datasets where it is typically required a pairwise computational approach where each column is compared to others. This work introduces a matrix-based algorithm that accelerates MI computation by leveraging vectorized operations and optimized matrix calculations. By transforming traditional pairwise computational approaches into bulk matrix operations, the proposed method enables efficient MI calculation across all variable pairs. Experimental results demonstrate significant performance improvements, with computation times reduced up to 50,000 times in the largest dataset using optimized implementations, particularly when utilizing hardware optimized frameworks. The approach promises to expand MI's applicability in data-driven research by overcoming previous computational limitations.
\end{abstract}

\section{Introduction}

Mutual Information (MI) is a foundational concept in information theory, introduced by Claude Shannon in his seminal 1948 work on communication theory~\cite{Shannon1948}. It quantifies the amount of shared information between two random variables, capturing both linear and non-linear dependencies. For binary data, MI is computed using the formula:

\begin{equation}
MI(X; Y) = \sum_{x \in \{0,1\}} \sum_{y \in \{0,1\}} p(x, y) \log_2 \left( \frac{p(x, y)}{p(x) p(y)} \right),
\end{equation}

where \( p(x, y) \) represents the joint probability of \( X \) and \( Y \) taking values \( x \) and \( y \) respectively, and \( p(x) \) and \( p(y) \) are the marginal probabilities of \( X \) and \( Y \). Unlike traditional correlation measures such as Pearson's correlation coefficient, which is limited to capturing linear relationships~\cite{Pearson1895}, MI is more versatile and can identify non-linear dependencies, making it particularly useful in high-dimensional and complex data settings.

The versatility of MI has led to its widespread adoption across numerous fields, including genomics~\cite{Reshef2011}, natural language processing~\cite{manning1999}, neuroscience \cite{Ince2017, Faskowitz2022}, and network science~\cite{Lizier2014, Tan2014}. It is extensively employed in feature selection and dependency assessment \cite{Vergara2014, Peng2005}, clustering \cite{Kraskov2005, Romano2014},, particularly in scenarios where traditional metrics may fail to uncover meaningful relationships \cite{Peng2005}. For example, MI has proven invaluable in selecting genetic markers associated with diseases \cite{Miotto2008} and in Computer security for intrusion detection \cite{Amiri2011, Sarem2021}.

Despite its theoretical importance and practical utility, the computational cost of MI remains a significant bottleneck when applied to large datasets, especially those consisting of binary variables. Computing pairwise MI for a dataset with \( n \) variables requires evaluating \( \binom{n}{2} \) pairs, which becomes infeasible as \( n \) grows. Existing algorithms typically employ pairwise calculations that are inherently sequential and lack the scalability required for high-throughput data analysis. This limitation has hindered the broader application of MI in fields where datasets are rapidly increasing in size and complexity.

The need for efficient computation is particularly acute in domains such as genomics, where binary data often arise from the presence or absence of mutations or gene expressions, and in network science, where binary adjacency matrices represent connections between nodes. Traditional computational strategies fail to take full advantage of modern hardware capabilities, such as vectorized operations, leaving a gap in the literature for fast and scalable MI computation methods.

In this paper, it is proposed a novel computation procedure designed to address these challenges. The approach suggested can be divided in two parts; the first one aims at vectorizing the algorithm by reducing it to simple matrix operations; the second part explores some of the potential computational redundancies to further optimize the whole procedure. By leveraging vectorized operations and optimizing for modern computational architectures, this method computes MI for all binary column pairs all at once, mostly with one single matrix multiplication. The proposed approach enables large-scale MI analyses that were previously computationally prohibitive, thereby broadening the applicability of MI to other research areas.

\section{Core algorithm presentation}

Traditional MI calculations involve summing over joint and marginal probabilities for each variable pair. However, this pairwise approach can be computationally expensive, especially for large datasets. To optimize this, the algorithm "unrolls" the summation of the MI formula and uses matrix operations to compute all required probabilities simultaneously.
The proposed algorithm leverages a matrix-based, bulk approach to compute mutual information (MI) across all pairs of binary variables in a dataset. Its essential idea is to use the Gram matrices of the data matrix for computing the counts of common values. The classical Gram matrix $D^T.D$ produces a matrix in which for each cell we get the counts of common ones for the corresponding row and column. For instance a Gram matrix with a value of 6 in position (3, 5), means that the number of ones in the same position for columns 3 and 5 of the data matrix is 6. This procedure allow us to compute only the common 1s, yet if we negate the original data matrix and compute similar Gram matrices with it, it becomes possible to count the ones that co-occur with zeros and also zeros that co-occur with zeros. These values can then be transformed into joint probabilities and marginal probabilities required and in one simple pass, compute all required elements for the application of expression (1).

The essential algorithm could then be enumerated like this
\begin{enumerate}
    \item Starting with an initial binary data matrix $D$, create the complementary matrix ($\neg D$);
    \item Compute the Joint Probability matrices using the Gram matrices from $D$ and $\neg D$;
    \item Use the diagonals of the matrices above for computing the Marginal Probabilities;
    \item Compute the expected values from the marginal probabilities;
    \item Combine all the elements above and sum them all to get all the pairwise values at once;
\end{enumerate}

These steps will now be explained in detail.

\newsec{1. Data Setup and Complementary Matrix}

Let \( D \) be an \( n \times m \) binary matrix, where each row represents a sample, and each column represents a variable. We construct a complementary matrix 

\[
 \neg D = 1 - D
\]
which will allow for efficient computation of joint occurrences for both values (0 and 1) across each variable pair.

\newsec{2. Joint Probability Matrices}

As referred above, the Gram matrix of the data \( G_{11} = D^T \cdot D \) displays the counts of each common 1s in each position of the matrix. Accordingly, we can compute all other possible counts using the complementary matrix \( \neg D = 1 - D \). Thus, we have:

\[
G_{00} = \neg D^T \cdot \neg D
\]

Similarly, for the cases where we have one variable being 1 and the other being 0, we compute the following:

\[
G_{01} = \neg D^T \cdot D
\]

and

\[
G_{10} = D^T \cdot \neg D
\]

With each of these, we can compute the joint probability matrices by dividing each Gram matrix by the total number of rows \( n \), thus:

\[
P_{11} = P(X = 1, Y = 1) = \frac{1}{n} G_{11}
\]

\[
P_{00} = P(X = 0, Y = 0) = \frac{1}{n} G_{00}
\]

\[
P_{01} = P(X = 0, Y = 1) = \frac{1}{n} G_{01}
\]

\[
P_{10} = P(X = 1, Y = 0) = \frac{1}{n} G_{10}
\]
Thus each $P$ matrix will account for all the joint probabilities probabilities of all column combinations

\newsec{3. Diagonal Elements for Marginal Probabilities}

The diagonal elements of \( G_{11} \) and \( G_{00} \) matrices provide marginal probabilities for each variable, namely  the probability vector for each variable being $1$:
    \[
    P_1 = P(X=1) = \frac{\text{Diag}(G_{11})}{n}
    \]

and the probability vector for each variable being $0$ is:
    \[
    P_0 = P(X=0) = \frac{\text{Diag}(G_{00})}{n}
    \]

\newsec{4. Bulk Computation of Expected Values Under Independence}

For the mutual information (MI) calculation, we need to compute the expected joint probabilities assuming independence between variables. These values are calculated using outer products of the marginal probabilities, thus

\[ E_{11} = P(X=1) \cdot P(Y=1) = P_1 \otimes P_1^T \]

\[ E_{00} = P(X=0) \cdot P(Y=0) = P_0 \otimes P_0^T \]

\[ E_{10} = P(X=1) \cdot P(Y=0) = P_1 \otimes P_0^T \]

\[ E_{01} = P(X=0) \cdot P(Y=1) = P_0 \otimes P_1^T \]

\newsec{5. Calculate Mutual Information for All Pairs}

To calculate the mutual information (MI) for binary variables \( X \) and \( Y \), we use equation (1), enrolling the summations, expanding it explicitly for all possible combinations of \( x \) and \( y \):

\begin{equation}
\begin{aligned}
\text{MI}(X, Y) &= P(X=1, Y=1) \log_2 \left( \frac{P(X=1, Y=1)}{P(X=1) \cdot P(Y=1)} \right) \\
&\quad + P(X=1, Y=0) \log_2 \left( \frac{P(X=1, Y=0)}{P(X=1) \cdot P(Y=0)} \right) \\
&\quad + P(X=0, Y=1) \log_2 \left( \frac{P(X=0, Y=1)}{P(X=0) \cdot P(Y=1)} \right) \\
&\quad + P(X=0, Y=0) \log_2 \left( \frac{P(X=0, Y=0)}{P(X=0) \cdot P(Y=0)} \right)
\end{aligned}
\end{equation}

As it is patent, we already computed all these variables for all variable pairs in matrix format, and we can then replace them in the final equation:

\begin{equation}
\begin{aligned}
\text{MI}(X, Y) &= P_{11} \log_2 \left( \frac{P_{11}}{E_{11}} \right) + P_{10} \log_2 \left( \frac{P_{10}}{E_{10}} \right) \\
&\quad + \, P_{01} \log_2 \left( \frac{P_{01}}{E_{01}} \right) + P_{00} \log_2 \left( \frac{P_{00}}{E_{00}} \right)
\end{aligned}
\end{equation}

we perform element-wise multiplication of each joint probability matrix with the logarithmic ratio of observed joint probabilities to expected probabilities under independence. For practical purposes, a small constant \( \epsilon \) is added to avoid division by zero within the logarithmic function, ensuring numerical stability.

This approach achieves high efficiency by leveraging matrix operations to compute joint and marginal probabilities across all variables in a single step, Using outer products for the denominator terms, which reduces the need for iterative pairwise computations and finally by avoiding explicit loops and pairwise comparisons through matrix operations, which are faster and more memory-efficient. By focusing on bulk matrix and broadcasting operations, the algorithm achieves substantial speedup, making it highly scalable for large binary datasets, making it particularly efficient in current-day GPUs. 

This efficient calculation of MI across all pairs of variables should be particularly useful for high-dimensional binary data, such as in genomics, text processing, and other applications with binary feature representations.

\section{Optimizing the computation of the Gram matrices}

The computation of \( G_{00}, G_{01} \) and \(G_{10}\) may pose computational challenges. Firstly because the computation of \( \neg D\) transforms a typically sparse matrix into a very dense matrix, minimizing the chances for taking advantage of sparse matrix computations, secondly because the Gram matrix computation (the most expensive part of the algorithm) is repeated three more times. Thus it would be important if some improvement could be made. And the basic idea is to identify what can we deduce from the Gram matrix computation involving  \( \neg D\)

\subsection{Simplified computation of \( \neg D\) Gram matrices}

We start with the definition of \( G_{00} \):

\begin{equation}
G_{00} = \neg D. \neg D  =(\mathbf{1} - D)^T \cdot (\mathbf{1} - D)
\end{equation}

where \( \mathbf{1} \) is an \( n \times m \) matrix filled with ones (assuming \( D \) is an \( n \times m \) matrix), and \( D \) is a binary matrix where each entry is either 0 or 1. Expanding this product, we get:

\begin{equation}
G_{00} = \mathbf{1}^T \cdot \mathbf{1} - \mathbf{1}^T \cdot D - D^T \cdot \mathbf{1} + D^T \cdot D
\end{equation}

Each of these terms will be analyzed in detail

\paragraph{Term 1: \(  \mathbf{1}^T \cdot \mathbf{1} \)}

Since \( \mathbf{1} \) is an \( n \times m \) matrix of ones, the product \( \mathbf{1}^T \cdot \mathbf{1} \) gives an \( n \times m \) matrix where each entry is the total number of samples \( n \):

\[
\mathbf{1}^T \cdot \mathbf{1} = \mathbf{N}
\]

Thus, we define the matrix \( N \) as an \( m \times m \) matrix where every entry has the value \( n \) and this can be immediately computed without any operation over the matrix, other than knowing the number of rows.

\paragraph{Term 2: \( \mathbf{1}^T \cdot D \)}

The term \( \mathbf{1}^T \cdot D \) results in an \( M \)-dimensional vector where each entry is the total count of ones in each column of \( D \). Let’s denote this vector by \( v \), where each element \( v_i \) is the sum of the entries in the \( i \)-th column of \( D \):

\[
v = D^T \cdot \mathbf{1}
\]

Using this vector, we can define a matrix \( C \) as an \( m \times m \) matrix in which each column is the vector \( v \), thus formally:

\[
\mathbf{C} =  \mathbf{1}_m \cdot \mathbf{v}
\]

where $1_m$ is a column vector of ones of size $m$. This matrix C can also be computed without any complex operation, only by counting the ones in matrix D (or individually summing each column), and replicating each produced row $m$ times to produce a squared $m \times m$ matrix

\paragraph{Term 3: \( D^T \cdot \mathbf{1} \)}

This is the transpose of Term 2, resulting in the matrix \( C^T \):

\[
D^T \cdot \mathbf{1} = C^T
\]

\paragraph{Term 4: \( D^T \cdot D \)}

This term is precisely the matrix \( G_{11} \), which gives the joint counts of ones for \( D \) itself.

\paragraph{Final Expression}

Putting it all together, we substitute each component back into the original expression for \( G_{00} \):

\begin{equation}
G_{00} = \mathbf{N} - \mathbf{C} - \mathbf{C}^T + G_{11}
\end{equation}

Now, as mentioned, matrices $N$ and $C$ can be readily computed without needing the large and dense $\mathbf{1}$ matrix, therefore avoiding further expensive matrix multiplications. Also both have a comparatively low $(m,m)$ dimension. Thus the $G_00$ matrix can be computed with sums and subtractions of components readily computed ($N$ and $C$ matrices) and already available ($G_{11}$)

This reasoning can be extended for the $G_{01}$ and $G_{10}$ matrices, and it can be verified that 

\begin{equation}
G_{01} = \neg D . D = (\mathbf{1} - D)^T \cdot  D = \mathbf{C} - G_{11}
\end{equation}

and obviously, as referred above

\[ G_{10} =  G_{01}^T \]

By computing  \( G_{00} \), \( G_{01} \) and \( G_{10} \) in this indirect manner, we save on computational resources, particularly avoiding further large-scale matrix operations. This approach ensures that we only perform one matrix multiplication for \( G_{11} \), with the remaining Gram matrices derived from simple element-wise matrix operations and transposition. This optimization is particularly advantageous for large or sparse datasets, making mutual information calculations feasible and efficient for high-dimensional binary data.

This method provides a significant speedup, particularly for large datasets with a high degree of sparsity, making it feasible to compute mutual information efficiently across all binary variable pairs.

\subsection{Algorithm complexity}

Despite the computation improvements, the overall time complexity of the algorithm is maintained as $O(m^2.n)$, being limited mostly by the computation of the Gram Matrix, where $D^T.D$ takes $m^2.n$ multiply-add operations. This is essentially the same complexity of the pairwise computation of  mutual information, where for all pairs of variables ($m^2$) the cross entropies are computed, going through all the values (thus $n$).  Nonetheless, the absence of explicit loops taking advantage of parallel implementations and hardware support for matrix operations should make the procedure even more efficient. 

\section{Implementation and results}

\subsection{Experimental setup}

The performance of different implementations for computing mutual information was evaluated using several frameworks and approaches. The basic algorithm was implemented using NumPy, which provides efficient array operations for dense matrices \cite{harris2020array}. To accelerate matrix multiplication in the computation of Gram matrices, the NumPy implementation was further optimized with Numba, a Just-In-Time (JIT) compiler that translates Python functions into optimized machine code \cite{lam2015numba}. Additionally, sparse matrices were handled using SciPy's sparse matrix functionality to improve memory efficiency for large, sparse datasets \cite{virtanen2020scipy}. The optimized algorithm was also tested using PyTorch  \cite{paszke2019pytorch}, which leverages tensor operations optimized for performanc. Also, despite the fact that PyTorch excels in using GPU resources, in this setup the built-in GPU was explicitly not used ensuring a fair comparison across implementations.

All experiments were conducted on a Mac Studio equipped with an Apple M2 CPU featuring 12 cores. As referred, the computations were carried out exclusively on the CPU, with no use of the GPU, to ensure consistency across the different methods. Matrix multiplication, a critical operation in the computation of mutual information, was central to the performance of each implementation. Both dense and sparse data representations were tested to explore the trade-offs between memory efficiency and computational overhead.

\subsection{Results and discussion}

\subsubsection{Global results}

The first set of tests involved running the pairwise approach with the basic algorith with and without the secondstage numeriacal optimizations. As the usage of numerical libraries is critical, it was also tested different implementations. The first one using NumPy and Numba, secondly, with scipy sparse matrices and thirdly with Pytorch. Three datasets of identical sparsity (90\%) but of different sizes were run through the implementations and their running times computed  (Figure \ref{tab:simple_results}).

\begin{table}[h!]
\caption{Running times (in seconds) for MI calculations across different implementations for 3 datasets with different numbers of rows and columns. (SKL Pairwise - Pairwise computation with Scikit learn Mutual Information; Bas-NN - Basic algorithm with NumPy and Numba; Opt-NN - Optimized Gram computation with NumPy and Numba; Opt-SS - Optimized Gram computation but using Scipy Sparse matrices; Opt-T; Optimized algorithm running with PyTorch}
\centering
\begin{tabular}{c c | r r r r r}
\toprule
\textbf{Rows} & \textbf{Cols} & \textbf{SKL Pairwise} & \textbf{Bas-NN} & \textbf{Opt-NN} & \textbf{Opt-SS} & \textbf{Opt-T} \\ 
\midrule
1000   & 100   & 1.430  &  0.001  & 0.001  & 0.001  & 0.021  \\ 
100000 & 100   & 54.389 & 0.064  & 0.013  & 0.033  & 0.061  \\ 
100000 & 1000  & 5,211.830 & 1.941 & 0.676 & 2.286 & 0.086 \\ 
\bottomrule
\end{tabular}

\label{tab:simple_results}
\end{table}

Computation times for the calculation of all mutual information (MI) values between columns of datasets of varying sizes have been produced (Table 1). For smaller datasets, the differences in computation time between approaches are less pronounced, but as dataset size increases, significant disparities emerge. The Scikit-learn pairwise computation is consistently the slowest across all sizes, with a computation time exceeding 5000 seconds for the largest dataset. The basic algorithm implemented using Numpy and Numba achieves drastic performance gains, computing the largest dataset in under 2 seconds. The Gram computation optimization to this algorithm yield even faster results,  reducing the computation times for the largest dataset almost 3 times. The sparse algorithm, utilizing Scipy, performs well in scenarios where sparsity can be exploited; however, for this experimental setup with approximately 90\% sparsity, it does not outperform the other approaches. Finally, the PyTorch implementation on CPU offers is clearly the fastest with computation times always below 0.1 secs, despite being slightly slower for the  smaller datasets due to initialization overhead. Overall, bulk computation methods are obviously superior and the PyTorch numerical superiority make them shine particularly well.

\subsubsection{Detailed implementation comparison}

As the pairwise implementation was so comparably inefficient, in the next batch of processing, it was decided to verify only the implementation effects, changing first the number of rows, considering always 1000 columns, and secondly, to inspect the effect of changing the number of columns, setting the number of rows fixed at 100,000 . 

\begin{figure}[h]
\includegraphics[width=12cm]{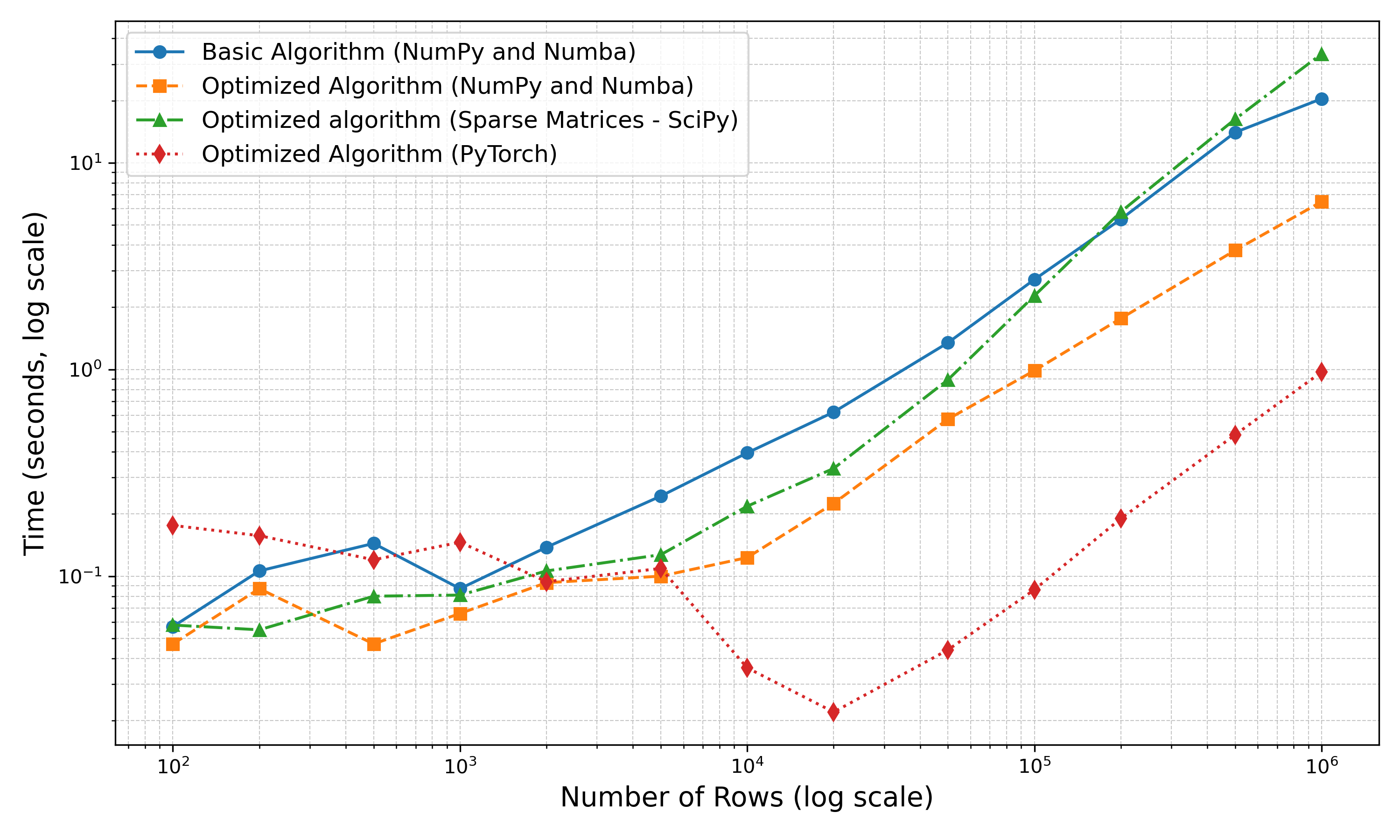}
\caption{Computation times of different algorithm implementations, changing the number of rows in the data set (number of columns fixed at 1000) }
\label{fig:performance-rows}
\end{figure}

Examining the impact of the dataset sizes on the algorithm performance (Figure 1) it can be verified that for datasets with fewer than 10,000 rows, all algorithms exhibit very low computation times, generally below 0.4 seconds, and the overhead of starting the procedure will dominate computation time. As the number of rows increases, the computation times grow but not linearly as the expected time complexity of the algorithm might suggest. 

Individually it can be seen the benefits of the second stage numerical optimization as the basic algorithm is generally the slowest approach. The usage of Sparse Matrices offer better performance than the dense matrix implementation but interestingly lags behind the optimized approach for larger datasets. This is consistent with the additional overhead of managing sparse data structures when sparsity isn’t fully leveraged. One further note relative to the Pytorch implementation as it exhibits a unique trend, with computation times dropping sharply after 20,000 rows (a consistent behavior verified over after several runs). Nonetheless this implementation is clearly the fastest, by orders of magnitude compared to the other approaches, even though the algorithm is the same. The reasons for this drop are probably hidden how Pytorch manages tensors and memory according to their sizes, possibly due to Dynamic Tensor Optimization, by internally leveraging optimized memory management techniques (e.g., caching or memory reuse) for larger tensors, leading to reduced overhead for operations on larger datasets, or distinct memory access patterns: On the testing platform (a Mac M2 architecture), specific memory access patterns for larger tensors may align better with the CPU’s architecture, improving performance. Our results seem to suggest that for datasets under 10,000 rows the differences between algorithms are negligible, making the choice of method less critical. For larger datasets, the optimized NumPy and PyTorch implementations are superior. PyTorch, in particular, demonstrates strong scalability for larger data sizes, making it an excellent choice for applications requiring processing of large datasets on modern hardware.

Changing the number of columns has a more notable impact as expected, due to the time complexity of the algorithm being squared on this factor.  

\begin{figure}[h]
\includegraphics[width=12cm]{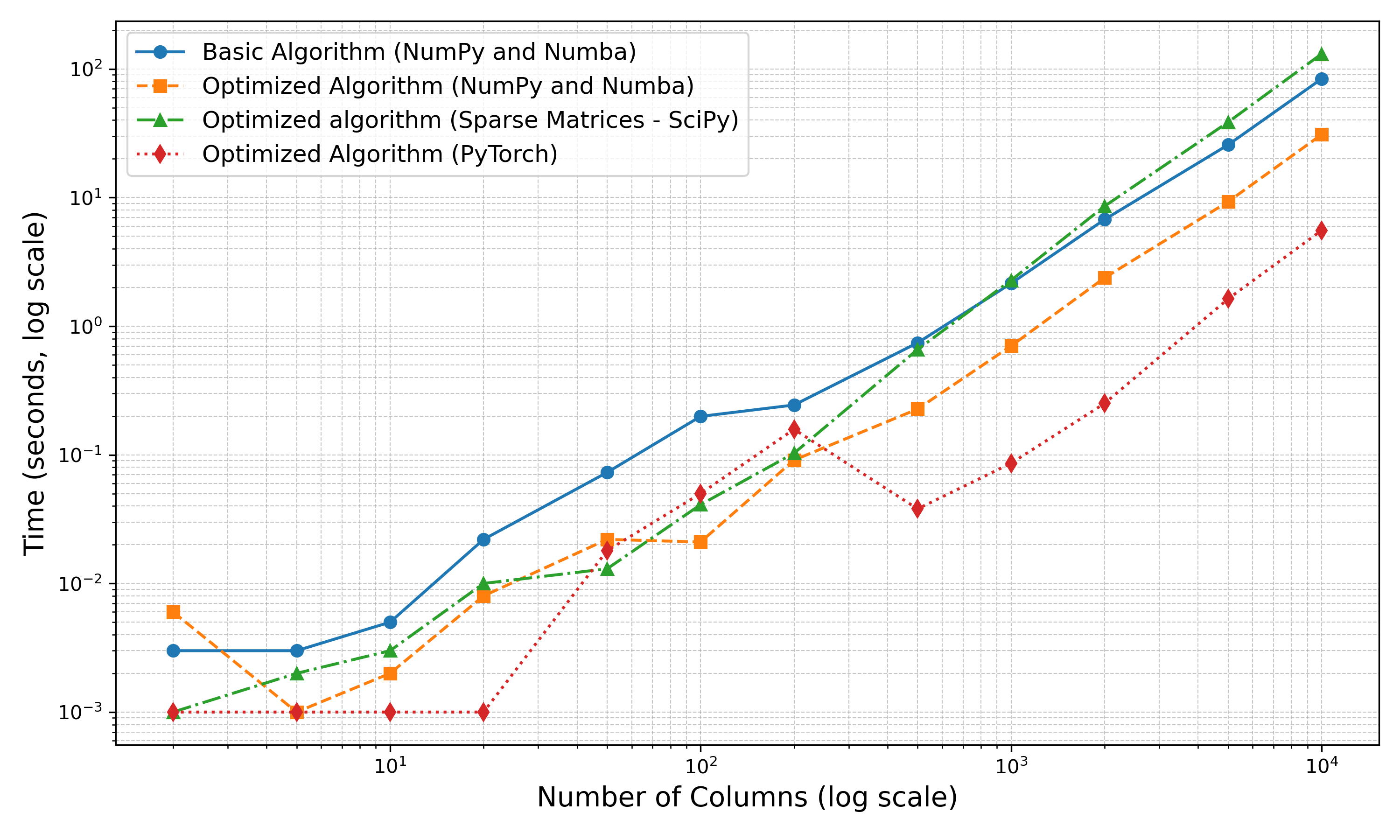}
\caption{Computation times of different algorithm implementations, changing the number of rows in the data set (number of columns fixed at 1000) }
\label{fig:performance-cols}
\end{figure}

The results presented (Figure 2)  show that the basic algorithm implemented using NumPy and Numba demonstrates a consistent and steep increase in computation time, reaching approximately 83.8 seconds on the largest examples. The optimized NumPy and Numba implementation, however, consistently outperforms the basic approach, maintaining a roughly threefold improvement in computation times across all tested dataset sizes. For example, at 10,000 columns, the optimized approach completes in 30.9 seconds compared to 83.8 seconds for the basic method, illustrating the efficiency gains from vectorization and algorithmic refinement.  The performance of the sparse matrix implementation using SciPy is less promising. While it performs comparably for smaller datasets, its efficiency diminishes significantly for larger datasets, taking approximately 130.9 seconds for 10,000 columns. This inefficiency may stem from the dataset's sparsity level (around 90\%), which is insufficient to fully leverage the computational benefits of sparse matrix operations, leading to overhead that outweighs its advantages at larger scales.  Finally, the PyTorch implementation, on the other hand, initially exhibits slower performance for datasets with fewer columns but scales exceptionally well for larger datasets. For 10,000 columns, it achieves a computation time of just 5.6 seconds, substantially outperforming all other approaches. This highlights the strengths of vectorized accelerated computation and parallel processing even on the CPU of which PyTorch is capable of harnessing.

\subsubsection{The effect of sparsity}

To actually understand the effect of sparsity on the computational times, one further essay was tried where similarly sized datasets of 100,000 rows and 1000 columns were generated, controlling only for the sparsity level, and the 4 optimized implementations were tested.

\begin{figure}[h]
\includegraphics[width=12cm]{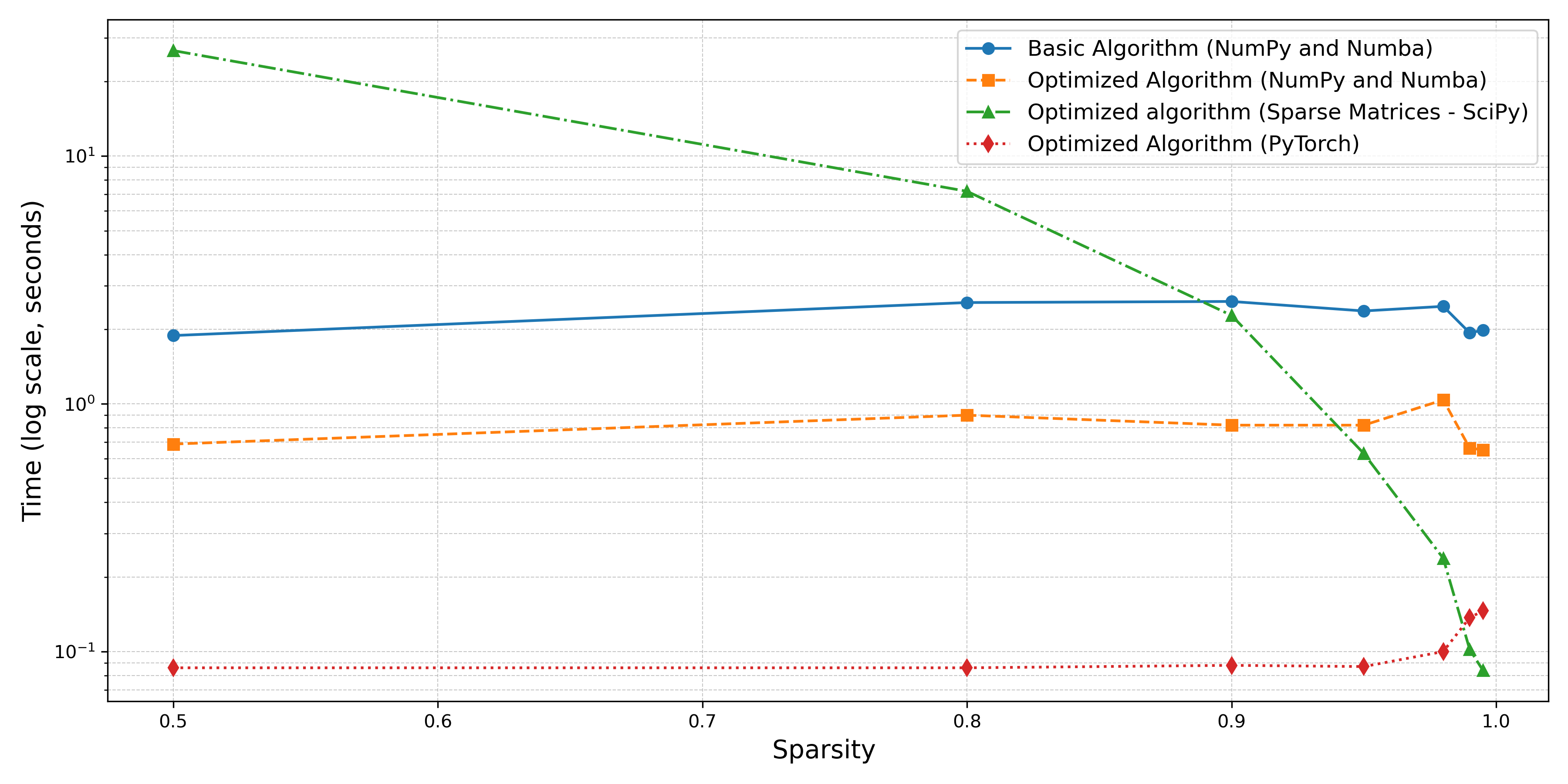}
\caption{Computation times (in seconds) across varying sparsity levels of the dataset (Simulated datasets with 100,000 rows and 1,000 columns) }
\label{fig:performance-sparse}
\end{figure}

The results (Figure 3) highlight the varying performance of the optimized methods as dataset sparsity changes. With the exception of the Sparse Scipy approach, the performance is similar across all sparsity levels. In contrast, the Scipy approach using sparse matrices benefits significantly from increasing sparsity, with computation times dropping over 300 times as the dataset goes from 50\% sparse, to 99.5\% , becoming, for the highest levels, the most efficient approach, faster than PyTorch and orders of magnitude faster than  the other approaches. It is also important to highlight that the Scipy sparse matrix approach was implemented by default and used  without the usage of any parallel processing methods for matrix multiplication, thus further improvements would be certainly possible. 

\section{Conclusions}
This research presents an optimized approach to mutual information computation for binary datasets. By reimagining MI calculation as a series of matrix operations rather than sequential pairwise computations, a new algorithm has been suggested that offers unprecedented computational efficiency. It was verified that the proposed method achieves substantial speedups across various dataset sizes, with computation times reduced by orders of magnitude compared to traditional approaches. Also, the optimization techniques used, including vectorization and intelligent matrix manipulation, enable rapid MI computation without compromising accuracy. Despite the fact that the algorithm's time complexity remains O(m²n), practical performance due to hardware optimization of vectorized calculations and  parallel matrix operations significantly outpaces traditional methods. It was further verified that matrix sparsity has no measurable effect on the overall computation, but if this is known beforehand, the usage of sparse matrix libraries can significantly boost performance. It was further verified that different numerical frameworks (NumPy, SciPy, PyTorch) demonstrate varied performance characteristics, with PyTorch showing exceptional scalability for large datasets even without direct GPU access 

The approach suggested is particularly effective for datasets with high dimensionality , making it valuable for emerging research domains with complex, high-dimensional binary data.

Future work could explore further optimizations, with blockwise computation for situations when the number of columns is too large, and its computation might exhaust the machine's memory, and extensions to non-binary datasets. By making more efficient mutual information computation, this method opens new possibilities for data-driven discovery across multiple scientific disciplines.

\section*{Source Code Implementation}
The Python source code for this work is available at the following GitHub Repository: \href{https://github.com/aofalcao/bulk-MI}{https://github.com/aofalcao/bulk-MI}

\bibliographystyle{plain}  
\bibliography{main}  

\end{document}